\documentclass[conference]{IEEEtran}
\usepackage[utf8]{inputenc}
\usepackage[backend=bibtex,sorting=none]{biblatex}

\usepackage{graphicx}
\usepackage{caption}
\usepackage{subcaption}
\usepackage{xcolor}
\title{FoodChem: A food-chemical relation extraction model }

\begin{document}

\author{\IEEEauthorblockN{1\textsuperscript{st} Gjorgjina Cenikj}
\IEEEauthorblockA{
\textit{Jožef Stefan International}\\
\textit{Postgraduate School}\\
Ljubljana, Slovenia \\ 
\textit{Computer Systems Department}\\
\textit{Jožef Stefan Institute}\\
Ljubljana, Slovenia \\
gjorgjina.cenikj@ijs.si
}

\and
\IEEEauthorblockN{2\textsuperscript{nd} Barbara Koroušić Seljak}
\IEEEauthorblockA{
\textit{Computer Systems Department}\\
\textit{Jožef Stefan Institute}\\
Ljubljana, Slovenia \\
barbara.korousic@ijs.si
}
\and
\IEEEauthorblockN{3\textsuperscript{rd} Tome Eftimov}
\IEEEauthorblockA{
\textit{Computer Systems Department}\\
\textit{Jožef Stefan Institute}\\
Ljubljana, Slovenia \\
tome.eftimov@ijs.si
}
}
\maketitle

\begin{abstract}
In this paper, we present FoodChem, a new Relation Extraction (RE) model for identifying chemicals present in the composition of food entities, based on textual information provided in biomedical peer-reviewed scientific literature. The RE task is treated as a binary classification problem, aimed at identifying whether the \textit{contains} relation exists between a food-chemical entity pair. This is accomplished by fine-tuning BERT, BioBERT and RoBERTa transformer models. For evaluation purposes, a novel dataset with annotated \textit{contains} relations in food-chemical entity pairs is generated, in a golden and silver version. The models are integrated into a voting scheme in order to produce the silver version of the dataset which we use for augmenting the individual models, while the manually annotated golden version is used for their evaluation. Out of the three evaluated models, the BioBERT model achieves the best results, with a macro averaged F1 score of 0.902 in the unbalanced augmentation setting.
\end{abstract}


\section{Introduction}
The proliferation of biomedical literature~\cite{ beyond_pubmed, publication_growth_coverage} has lead to an endless amount of electronic health records and more than 30 million citations and abstracts in PubMed~\cite{pubmed}. Numerous studies have been conducted which investigate the relations between food consumption and the development of different diseases such as obesity~\cite{hebebrand2021concept}, cardiovascular disease~\cite{zhong2021association}, etc. Linking food consumption data to food composition data is crucial for conducting such studies, however, the missing values present in food composition databases~\cite{merchant2006food} pose a limitation on the extensiveness of the conducted research, and introduce additional challenges with microbioma studies~\cite{annalisa2014gut}. To overcome these issues, a lot of research has already been conducted on the topic of identification of the chemical composition of different food entities. Consequently, a large quantity of valuable results and research outcomes are available as unstructured text data in scientific papers. 

In order to be able to utilize and structure this knowledge, we need to focus on pipelines that are able to automatically extract and recognize the relations between foods and chemicals. Manual structuring of this information is characterized with high precision and reliability~\cite{curration_accuracy}, but it requires the availability of human annotators, whose inability to keep up with the overwhelming amount of research~\cite{knowlife}, results in lower recall rates, poor scalability and time efficiency~\cite{biomed_kgc_min_supervision}.

In order to introduce automation into this process, Information Extraction (IE) pipelines consisting of Named Entity Recognition (NER), Named Entity Linking (NEL) methods, and Relation Extraction (RE) methods can be used. NER methods extract mentions of specific types of entities from raw text. NEL methods further link the extracted entities to concepts in different knowledge bases, and finally, RE methods detect relations between the entities~\cite{portugese_re_sequence_model}. The outputs of such IE pipelines can then be used to automatically suggest relations between concepts in different knowledge bases to which the NEL methods can link the entities to. In semi-automatic approaches, the suggested relations are examined and corrected by domain experts. The goal of such methods is to decrease the amount of effort required by experts while maintaining the precision of manual annotation and increasing the recall rates, to efficiently handle the rising publication rate and eliminate the existing annotation bottleneck.

The development of such IE pipelines in the domain of food and nutrition is largely impeded by the lack of annotated data which can be used for training the NER and RE models. In this paper, we introduce a small golden corpus of sentences annotated with respect to the existence of a \textit{contains} relation between food and chemical entities. We further train RE models to identify chemicals present in food entities by mining biomedical scientific literature. We perform fine-tuning of the transformer-based models BERT~\cite{bert}, BioBERT~\cite{biobert} and RoBERTa~\cite{roberta} with a small amount of manually annotated data, and use a voting scheme of the three models to automatically annotate additional data, which is used for augmenting the models. Our focus is not to generate an ensemble RE model, but rather, to create a silver corpus of automatically annotated samples and evaluate the impact of augmenting the individual models with the silver corpus. Both the golden and the silver corpus are publicly available on Github\footnote{https://github.com/gjorgjinac/food\_chem}.

\section{Related work}
\subsection{Relation extraction in the domain of food and nutrition}
Automating the RE task has been the goal of numerous works in the biomedical domain in the past decade. Several methods have been developed for extracting relations between chemicals and proteins~\cite{cpi_re2,cpi_re1,cpi_re3}, chemicals and diseases~\cite{ cd_re2, cd_re, cd_re3} and chemicals and genes~\cite{cg_re1}.

In the domain of food and nutrition, some research efforts have been dedicated towards the extraction of food relations for commercial applications in the German language~\cite{ food_re_wo_training_data, data_driven_food_ie, german_re} and extracting food-disease relations from scientific literature for the Chinese language~\cite{chinese_re}. In 2021, FoodMine has been presented~\cite{hooton2020exploring}, which is a pipeline that can classify research abstracts into three classes (i.e., \textit{not useful}, \textit{quantified}, and \textit{unquantified}) depending on the information of food chemical composition presented in the text. FoodMine involves the training of a classification model using a manually annotated corpus of abstracts related only to cocoa and garlic. The main limitation is that it can detect an abstract where food chemical composition information is presented, however, the concrete information needs to be extracted manually by experts.

For the English language, RE methods exist for detecting relations between food and disease entities~\cite{saffron,dietrx,nutrichem2}. We have previously developed the SAFFRON~\cite{saffron} RE model, where we propose the use of transfer learning (TL) to make up for the lack of annotated data with respect to relations between food and disease entities. SAFFRON's methodology bears some similarities to the one presented in the current work, however, while SAFFRON extracts relations between food and disease entities, this work is focused on food-chemical relations. Another related resource is NutriChem~\cite{nutrichem2}, which is, to the best of our knowledge, the only resource which contains automatically extracted food-chemical relations. However, its focus is on plant-based foods only, while we target a complete range of food categories.

\subsection{Text representation models}
In the past few years, state-of-the-art results in several natural language processing tasks have been achieved using transformer models~\cite{nlp_sota}. In our experiments, we use three transformed-based models (BERT, RoBERTa and BioBERT), which follow the principles of transductive TL, meaning that they are pre-trained on large quantities of data and can then be fine-tuned to perform other downstream tasks such as NER, RE, Natural Language Inference or Question-Answering (QA). This can be accomplished without substantial changes in the original architecture. In the simplest case, only the output layer needs to be modified to suit the downstream task.

The original BERT (Bidirectional Encoder Representations from Transformers)~\cite{bert} model is a bidirectional, contextual representation model pre-trained on an unsupervised Mask Language Modeling or Next Sentence Prediction task.
We use the original BERT model, which is pre-trained on the BooksCorpus~\cite{books_corpus} and English Wikipedia.

The RoBERTa (Robustly Optimized BERT Approach)~\cite{roberta} model is based on the original BERT architecture, with several adjustments in the pre-training phase, such as increasing the training time, training on a larger amount of data, removal of the Next Sentence Prediction task, and introduction of dynamic masking. Apart from the datasets used for pre-training the original BERT model, 3 additional sources are involved in the pre-training of RoBERTa: the OpenWebText corpus~\cite{open_web_text_corpus}, the Stories subset from the Common Crawl dataset~\cite{common_crawl_stories} and the CommonCrawl News dataset~\cite{ccnews}.

While the BERT and RoBERTa models are not domain-specific, the BioBERT (Bidirectional Encoder Representations from Transformers for Biomedical Text Mining)~\cite{biobert} model is intended for application in the domain of biomedicine. The data on which BioBERT is trained is supplemented by PubMed abstracts and full-text articles from PubMed Central, and thus, BioBERT has been shown to outperform BERT in biomedical NER, RE, and QA ~\cite{biobert}.

\section{Methodology}
\begin{figure}
    \centering
    \includegraphics[width=\linewidth]{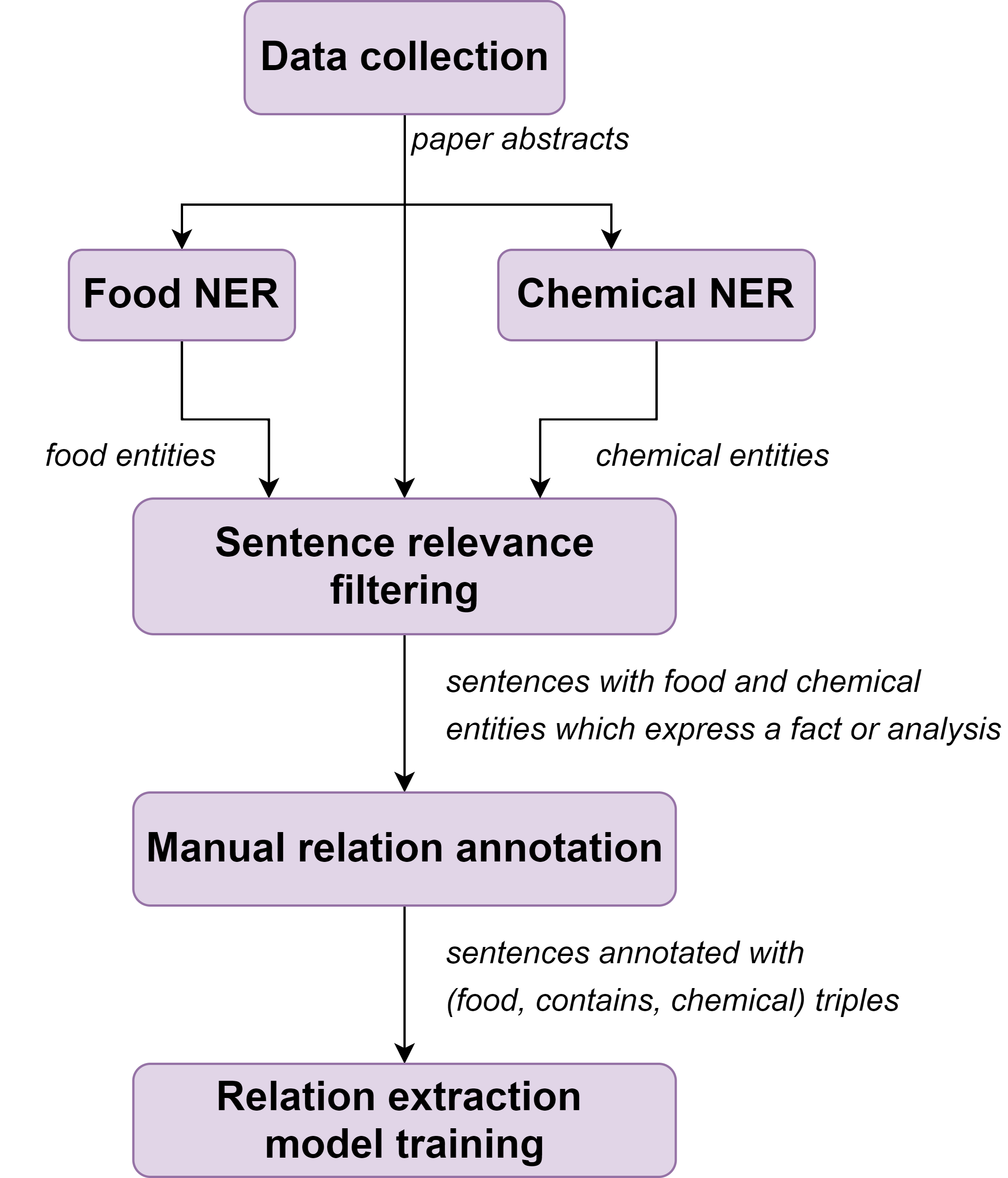}
    \caption{Overview of the pipeline used for annotating scientific abstracts}
    \label{fig:annotation_pipeline}
\end{figure}

The overview of the proposed pipeline for annotating the data needed for training the RE models is featured in Fig.~\ref{fig:annotation_pipeline}. The initial step entails retrieving abstracts of peer-reviewed scientific papers from PubMed. Next, NER and NEL methods (described in more details in the following subsections) are applied for identifying the chemical and food mentions in the text of the abstracts. The sentences which express a fact or analysis of the obtained result, and contain at least one food and one chemical entity, are extracted from the text of the abstracts. These sentences are manually annotated for the existence of a \textit{contains} relation between the food-chemical entity pairs, and the annotations are used for training a RE method, which determines whether a sentence expresses the fact that the chemical entity is contained in the food entity.
In the following subsections, we explain each step in more detail.

\subsection{Data collection}
100,000 paper abstracts are collected from PubMed using the Entrez Programming Utilities~\cite{entrez}, where the term \textit{food chemical} was used as a search keyword, in order to extract relevant papers. Apart from the text of the abstract, the following information was also extracted for each paper: the paper title, the year the paper was published, the journal it was published in, and the MESH terms related to it.

\subsection{Chemical Named Entity Recognition}
In order to extract the chemicals from the raw text of the abstracts, we use the Sequence Annotator for Biomedical Entities and Relations (SABER)~\cite{saber}. SABER is a biomedical NER tool, which combines several strategies for improving the generalization ability of the commonly used NER architecture based on Bidirectional Long Short-Term Memory and Conditional Random Fields (BiLSTM-CRF). Multi-task learning is applied to jointly train the hidden layers with different corpora, variational dropout is used for increasing the regularization of the recurrent layers, and TL is implemented by starting the training with a large, semi-automatically generated silver corpus, and continuing the training with a golden corpus~\cite{saber}.

The tool provides several pre-trained models for the recognition of disorders, chemicals, organisms, genes and gene products.
We use the \textit{CHED} pre-trained model~\cite{saber} which is capable of identifying chemical entities using common and trademark names, abbreviations, molecular formulas, chemical database identifiers and names defined in the nomenclature of the International Union of Pure and Applied Chemistry. We broadly refer to all of these as chemicals. Apart from localizing the chemical mention in the text, SABER is also capable of linking the extracted entities to the PubChem database~\cite{pubchem}.

\subsection{Food Named Entity Recognition}

BuTTER~\cite{butter} is a food NER method trained on the golden version of the FoodBase corpus~\cite{FoodBase}, which contains 1000 recipes annotated for the existence of food entities. Much like the SABER model, BuTTER also uses the BiLSTM-CRF architecture, and evaluates the impact of supplementing the word embeddings of different word representation models with additional character embeddings. We use the lexical lemmatized BiLSTM-CRF model without character embeddings, which achieved the best results in terms of the averaged macro F1 score in the original work~\cite{butter}.
However, since the BuTTER model is trained on texts from recipes, it fails to generalize well on scientific text, which is observable from the increased amount of false positive entities extracted by the model. For this reason, we introduce two additional dictionary-based NER methods which use the names of food entities in the FooDB database~\cite{foodb}.
The FooDB database contains common and scientific names of foods, descriptions, macronutrient and micronutrient information, including chemicals that give foods their taste, aroma, texture and color. Each chemical is described by more than 100 compositional, biochemical and physiological attributes. The food concepts are further linked to the Integrated Taxonomic Information System (ITIS)~\cite{itis}, Wikipedia~\cite{wikipedia} articles and concepts in the National Center for Biotechnology Information Taxonomy (NCBIT)~\cite{ncbi}. The food group and subgroup are also retrieved from FooDB.

We implement two dictionary-based models, which use the names of food concepts defined in FooDB and perform simple string matching to extract the food mentions from the text of the abstracts. The dictionary of the FooDB non-scientific method consists of the common names of 992 foods, while the dictionary of the FooDB scientific method contains 675 scientific names of foods available in the FooDB database.

To increase the precision of the BuTTER model, we combine its annotations with the annotations of the FooDB non-scientific model, and consider the extracted food entities to be valid only if they were extracted by both of the models. Since BuTTER is not trained to recognize entities using their scientific names, we cannot combine it with the FooDB scientific model, so we consider all entities extracted by the FooDB scientific model to be valid.

\subsection{Sentence relevance filtering}
The pipeline extracts relations on a sentence level.
The sentence relevance filtering (SRF) model is tasked at determining which of the sentences in the abstracts are relevant for extracting relations.

As a preliminary step, we filter out the sentences which do not contain at least one food and one chemical entity. 

Abstracts typically summarize the full text of the paper, stating the topic, objective, hypothesis, methodology, and main findings of the research.
However, not all of these pieces of information are a reliable source for extracting relations. For instance, if the abstract contains a sentence describing the authors' hypothesis, which was ruled out as incorrect by the results, and we were to extract information from the sentence that describes that hypothesis, our findings would be based on incorrect information. For this reason, it is necessary to distinguish these sentences and filter out the irrelevant ones. 
We define a relevant sentence as one that contains at least one pair of food and chemical entities, and expresses a previously proven fact or a finding of the current study. 

In our previous work, we have developed a model that is able to make this distinction, and is based on the assumption that a single sentence is either relevant or irrelevant. The SRF model was trained on the GENIA Meta-knowledge event corpus~\cite{genia_enriched}, which is a collection of MEDLINE articles, annotated with meta-knowledge annotations, such as the general information type of the event (whether it is a fact, experimental result, investigation, etc), the level of certainty, the polarity of the event (positive or negative), etc. 

The model predicts a binary label derived from the \textit{Knowledge Type} annotation in the GENIA Meta-knowledge event corpus, which can take one of the following values: \textit{Investigation}, \textit{Analysis}, \textit{Method}, \textit{Fact}, \textit{Other}.
Even though the corpus is annotated with events, the annotations related to events are disregarded, because the model works at a sentence level. Since a sentence might contain several events, for the training of the model, we extracted unique, individual sentences that have the \textit{Knowledge Type} annotation, and only use the raw text, without any event annotations.

The SRF model is a binary classifier which distinguishes between relevant sentences (ones that are annotated with \textit{Analysis} or \textit{Fact}) and irrelevant sentences (annotated with \textit{Investigation} or \textit{Method}). 
The SRF classifier is based on end-to-end fine-tuning of a BERT model, with the last layer of the model replaced with a dropout and a linear layer which performs binary classification. 
The evaluation of the SRF model using stratified 10-fold cross validation, produced a macro averaged F1 score of 0.81, and an averaged precision of the positive class of 0.90. The model used in practice is trained on 90\% of the whole data, with the remaining 10\% being used for validation.

\subsection{Relation extraction}
A subset of 493 extracted sentences are manually assigned a binary label indicating whether a \textit{contains} relation exists between the two entities, i.e., whether the sentence indicates that the chemical entity is a component of the food entity. Out of these sentences, 327 express the \textit{contains} relation between the two entities, meaning that the dataset is unbalanced, with 327 positive and 166 negative samples. We refer to these manually annotated sentences as the golden corpus.

 We use these annotations to train binary classifiers which learn to recognize the \textit{contains} relation by fine-tuning the pretrained transformer models BERT~\cite{bert}, BioBERT~\cite{biobert} and RoBERTa~\cite{roberta}. The entities occurrence in each sentence is masked out in order to mark the entities' position in the sentence, i.e. food entities are replaced by \textit{XXX}, and chemical entities are replaced by \textit{YYY}. A similar approach has been used in~\cite{saffron}. This preprocessing has several objectives: to enable the classifiers to learn relations from complex sentences which have several food-chemical pairs, to prevent the classifiers from learning relations between concrete instances of food-chemical pairs, and instead encourage the classifiers to learn from the context words used to describe the sentence. 

\begin{figure}
    \centering
    \includegraphics[width=\linewidth]{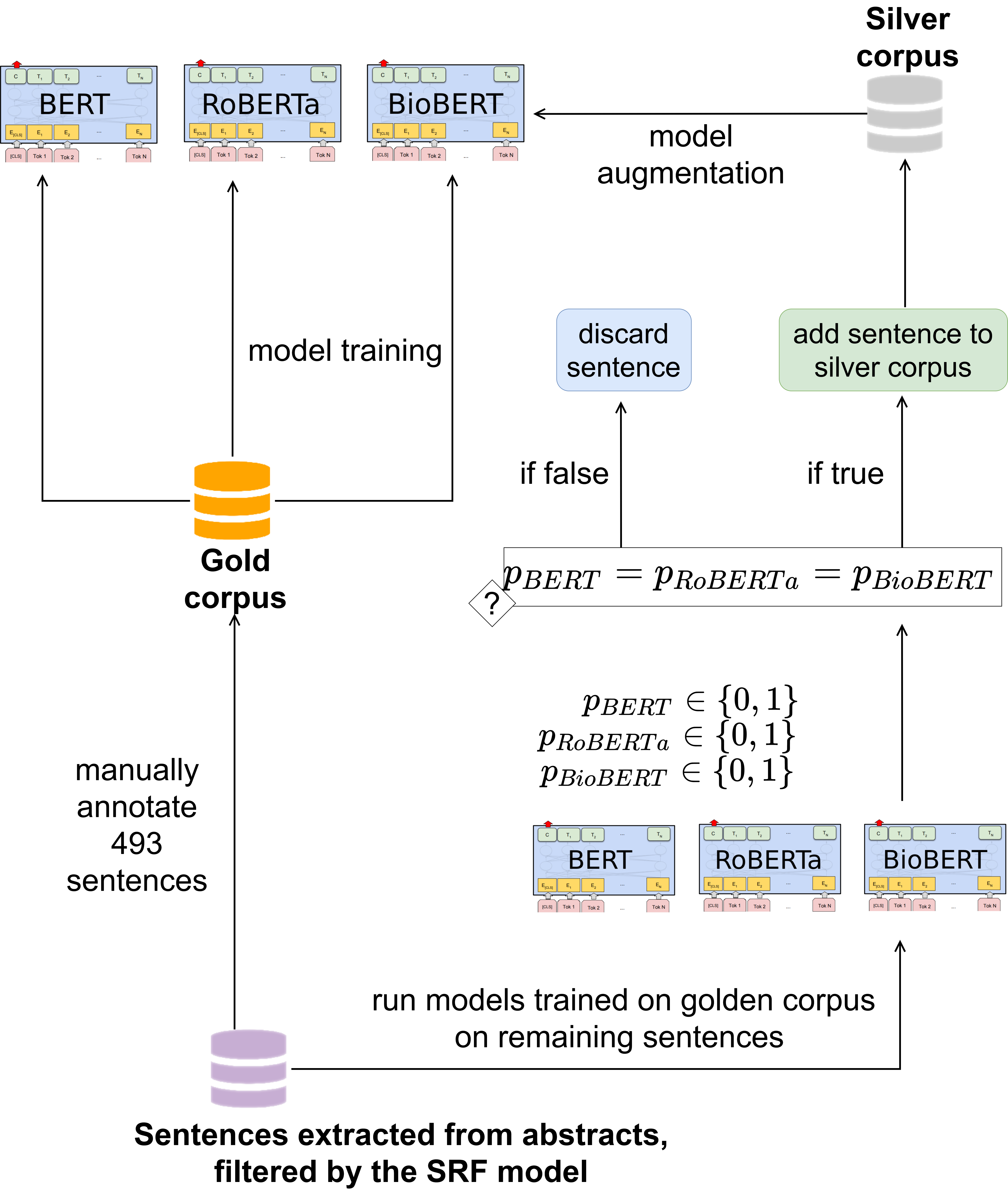}
    \caption{Training of the RE models and annotation of the silver corpus}
    \label{fig:training_pipeline}
\end{figure}

We apply each of the three models to the rest of the extracted sentences, which were not manually annotated.
We apply a voting scheme, which assigns a positive label for the existence of a \textit{contains} relation if all three of the models produce a positive label, and a negative label if all of the models produce a negative label. If the three models produce conflicting labels, we discard the sentence. By applying the voting scheme, we annotate a silver corpus consisting of 1,396 positive and 3,949 negative sentences. Fig.~\ref{fig:training_pipeline} depicts this process visually.

Using both the sentences in the silver corpus and the manually annotated ones, we retrain each of the three models.
All things considered, we train the models using three strategies: 
\begin{itemize}
    \item non-augmented - This is the initial strategy, where only the golden corpus is used for training. The evaluation is done using stratified 10-fold cross validation, so the training set contains 294 positive and 149 negative samples.
    \item augmented-unbalanced - In this strategy, all of the instances from the silver corpus are added to the training portion of the golden corpus, and are used to train the model. This results in an unbalanced class distribution, with 1,690 positive samples, and 4,098 negative samples.
    \item augmented-balanced - In this strategy, instances from the silver corpus are again added to the training portion of the golden corpus, however, the excess negative samples that are present in the augmented-unbalanced strategy are omitted using random sampling, to obtain a balanced distribution with 1,690 positive and 1,690 negative samples.
\end{itemize}
In each strategy, the same testing and validation sets from the golden corpus are used. For each fold, 30\% of the samples are removed from the testing set and used for validation. 
\section{Results}
\subsection{Experimental setting}
The transformer models are trained on the Google Colab~\cite{google_colab} platform, using the simpletransformers\footnote{https://simpletransformers.ai/} python library.
 The last layer of each of the transformer models (BERT, RoBERTa and BioBERT) used for the SRF and RE tasks is replaced with a dropout and a linear layer to perform binary classification. During fine-tuning, the model parameters are initialized with the values from the pre-training step, and are fine-tuned for the binary classification task.
The models are trained for a maximum of 10 epochs, using the AdamW optimizer with a learning rate of $4*10^{-5}$.
In order to prevent overfitting, an early stopping strategy is used, meaning that the training is interrupted once the decrease in the evaluation loss does not exceed the threshold of $5*10^{-3}$ for 2 consecutive epochs.
The source code for training the RE models is available at Github~\footnote{https://github.com/gjorgjinac/food\_chem}.
\subsection{Dataset generation}
Table~\ref{tab:ner_stats_table} features the number of entity mentions which were extracted by each of the NER methods from the texts of the 100,000 paper abstracts retrieved from PubMed.

\begin{table}[]
    \caption{Number of food and chemical mentions extracted by each of the NER methods}
    \centering
    \begin{tabular}{|p{95px}|p{60px}|p{60px}|}
    \hline
    Model name& Total entity 
    
    mentions &Unique entity 
    
    mentions
         \\\hline
         BuTTER (food) & 1,677,840 & 341,441 
         \\\hline
         FooDB non-scientific (food)& 59,704 & 1,088 
         \\\hline
         FooDB scientific (food) & 7,515 & 523 
         \\\hline
         Food Voting Scheme (food) & 26,315 & 880 
         \\\hline
         SABER (chemical) & 587,039 & 89,748
         \\\hline
    \end{tabular}
    \label{tab:ner_stats_table}
\end{table}

We use the food entities extracted by the Food Voting Scheme, and the chemical entities extracted by the SABER model to find sentences which contain at least one food and one chemical entity. We identify 24,515 such sentences, 8,068 of which are assigned a positive label by the SRF model, meaning that they express a \textit{Fact} or \textit{Analysis}.

The golden version of the dataset that is manually labeled for the existence of \textit{contains} relations between food-chemical pairs contains 493 labeled sentences, 327 positive and 166 negative, while the silver corpus contains 1,396 positive and 3,949 negative sentences. Each sample consists of the name of the two entities, the sentence in which they co-occur, their position in the sentence, the identifier of the chemical entity in PubChem, the group and subgroup of the food entity, and the identifiers of the food entity in FooDB, ITIS, Wikipedia, and NCBIT.

\subsection{Relation Extraction}
\begin{figure}
     \centering
     \begin{subfigure}[b]{\linewidth}
         \centering
         \includegraphics[width=\linewidth]{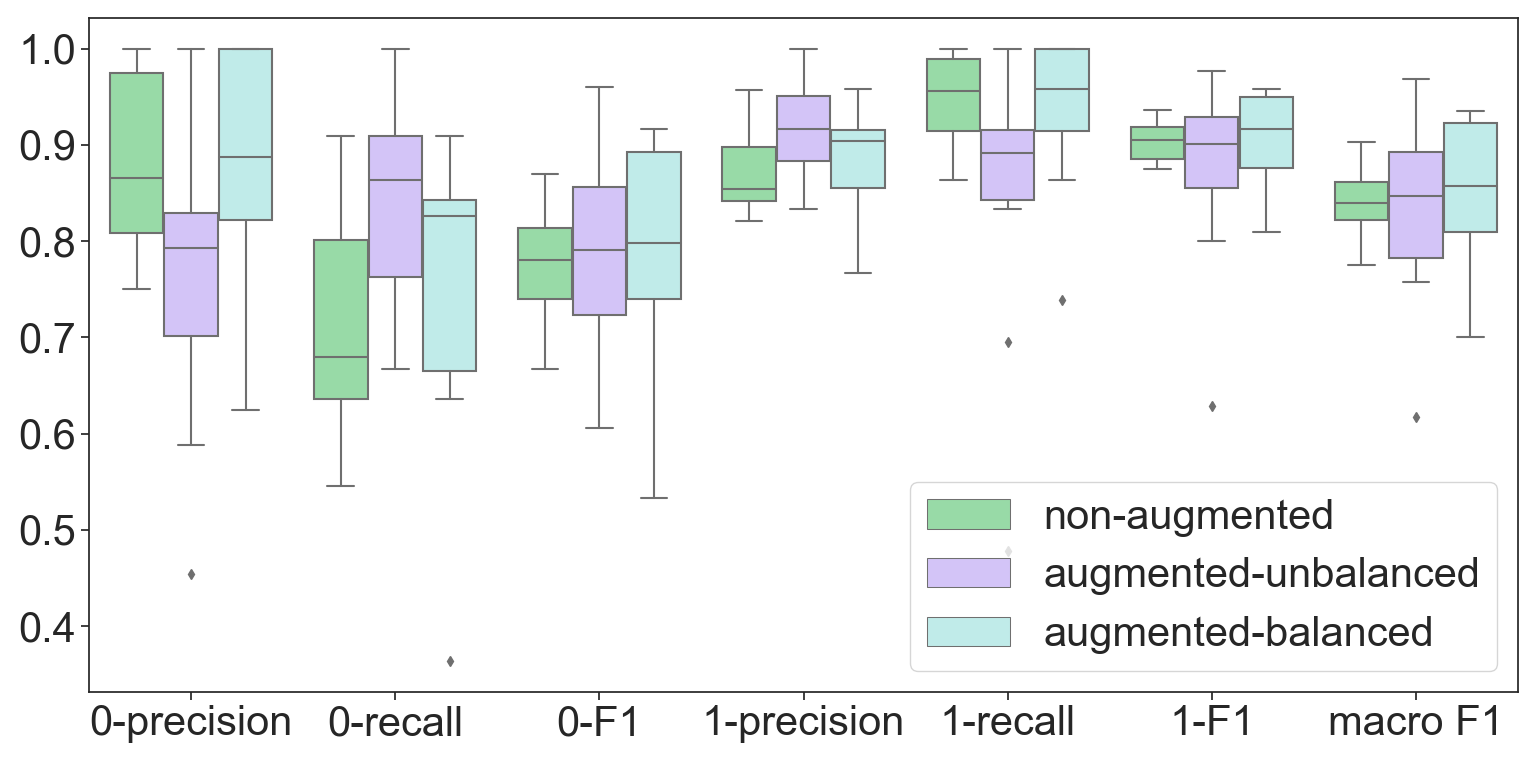}
         \caption{Results from the evaluation of the BERT model}
         \label{fig:results_bert}
     \end{subfigure}

     \begin{subfigure}[b]{\linewidth}
         \centering
         \includegraphics[width=\linewidth]{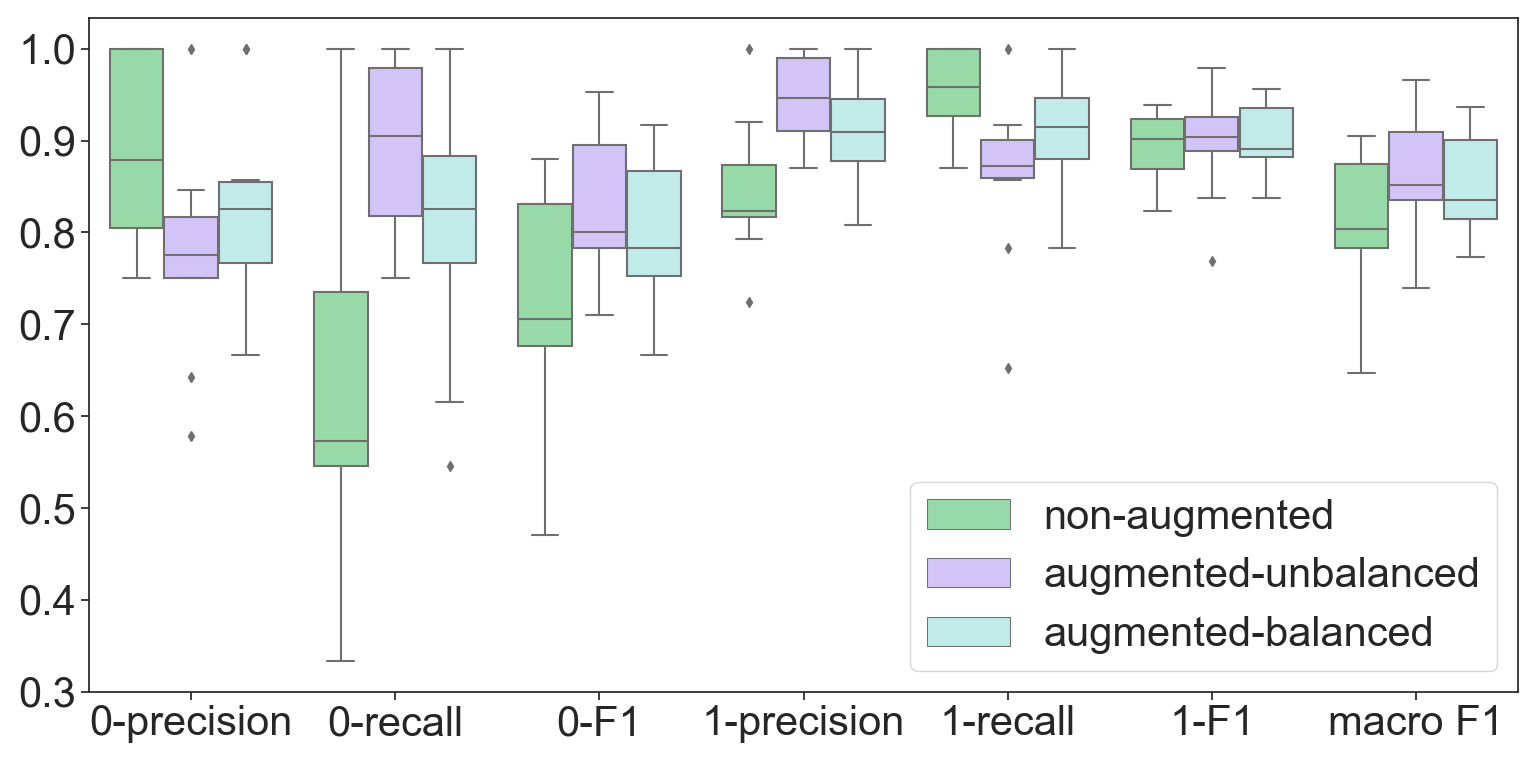}
         \caption{Results from the evaluation of the RoBERTa model}
         \label{fig:results_roberta}
     \end{subfigure}

     \begin{subfigure}[b]{\linewidth}
         \centering
         \includegraphics[width=\linewidth]{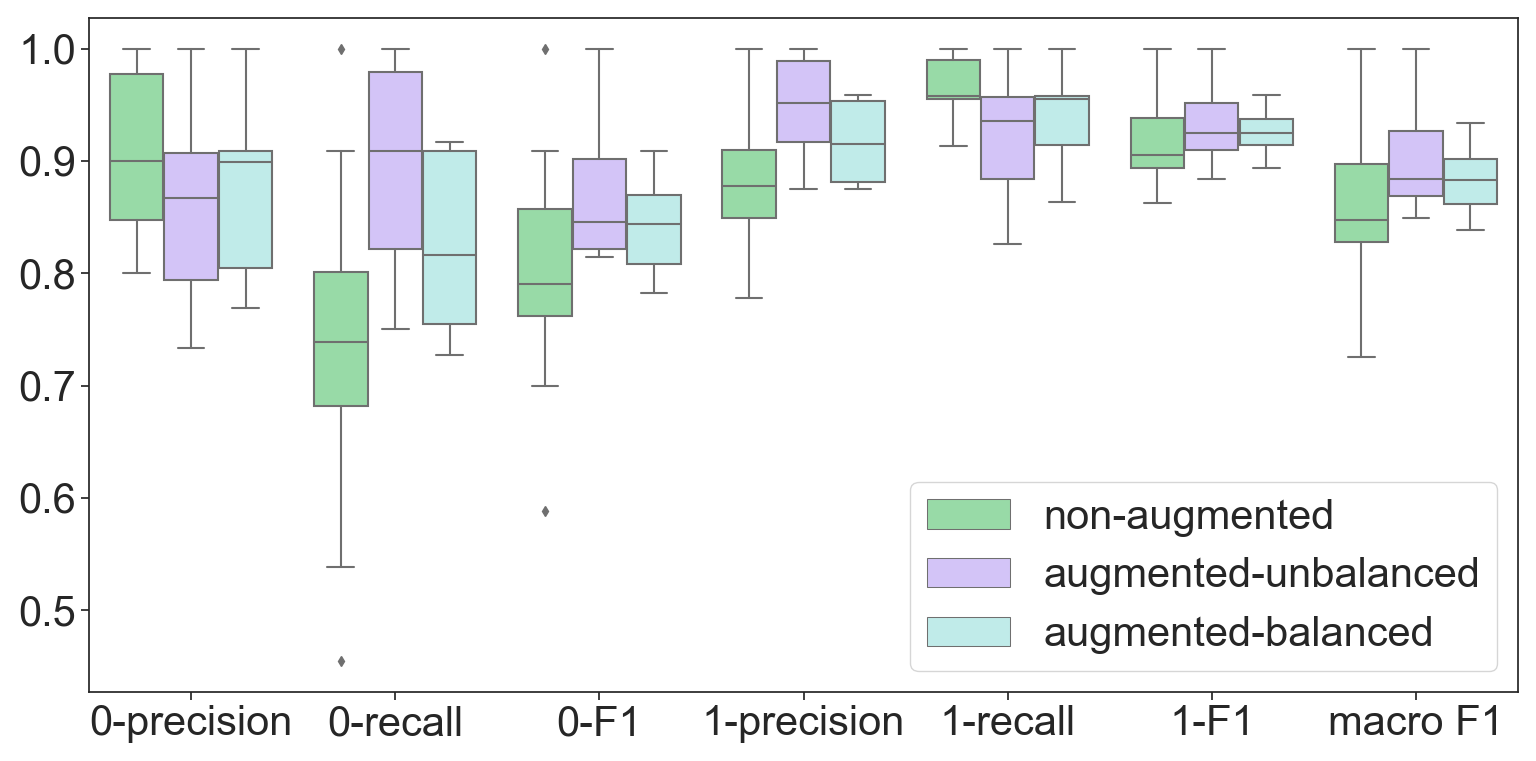}
         \caption{Results from the evaluation of the BioBERT model}
         \label{fig:results_biobert}
     \end{subfigure}
        \caption{Results from the evaluation of the RE models}
        \label{fig:results_re}
\end{figure}

The performance of the BERT, RoBERTa and BioBERT models is evaluated using stratified 10-fold cross validation, and the results are depicted in the form of boxplots in Fig.~\ref{fig:results_re}. For each model, we report the precision, recall and f1-score of the positive (1) and negative (0) class, and the macro averaged f1-score. These are depicted on the x-axis in the subfigure for each model. The green plots refer to the evaluation results of the non-augmented models, the purples plots refer to the results of the augmented-unbalanced models, while the blue plots refer to the results of the augmented-balanced models.

The models trained with the same strategy share a few common patterns. 
For most of the models, the augmentation (both balanced and unbalanced) results in an increase off all of the metrics except for the precision of the negative class and the recall of the positive class, compared to the non-augmented models. This is likely a consequence of the class distribution in the two datasets: the positive class is dominant in the training sets of the non-augmented models, which could lead to the models predicting the positive class more often than the negative class, thus having higher precision rates for the negative class, and higher recall rates for the positive class, which is not the case for the augmented models, where the negative class is dominant (in the augmented-unbalanced models) or the distribution is uniform (in the augmented-balanced models). 

Compared to the augmented-unbalanced models, in the augmented-balanced models, we observe an increase in the precision of the negative class and a decrease in its recall, a decrease in the precision of the positive class, and an increase in its recall. Again, this makes sense when the class distribution is considered.

In terms of the macro averaged f1-score, averaged from all folds, the BioBERT model achieves the best results, with a score of 0.859 for the non-augmented model, 0.902 for the augmented-unbalanced model, and 0.885 for the augmented-balanced model. The RoBERTa model achieves scores of 0.808, 0.861, and 0.853 for the non-augmented, augmented-unbalanced and augmented-balanced model, respectively, while the BERT model achieves scores of 0.839, 0.829 and 0.850. According to these results, we can conclude that the augmentation generally improves the performance, the only exception being the non-balanced augmentation of the BERT model, which experiences a reduction in the average macro averaged f1-score of 0.010, compared to the non-augmented model.

Table~\ref{tab:re_examples_table} features several examples of relations that were identified by our method. In the fourth sentence, we observe a partially extracted food entity, where instead of extracting the full \textit{fruits of Forsythia suspense} entity, the food NER method only captures the word \textit{fruits}, since the term {Forsythia suspense} is not present in the FooDB database. Such partial matches can also be extracted by the disease NER method, as is the case with the word \textit{phenolic} being captured in the third example, instead of \textit{phenolic antioxidant compound}.

\begin{table}
    \caption{Examples of relations extracted by the model}
    \centering
    \begin{tabular}{|p{100px}|p{130px}|}
    \hline
   Sentence & Extracted relations 
    \\\hline
     1-O-(4-hydroxymethylphenyl)-Î±-L-rhamnopyranoside (MPG) is a phenolic glycoside that exists in Moringa oleifera seeds with various health benefits, whereas its hepatoprotective effect is lacking clarification.
      &
     \textbf{Food}: moringa oleifera
     \newline
    \textbf{Chemical}: 1-O-(4-hydroxymethylphenyl)-Î±-L-rhamnopyranoside
   \newline
    
   \textbf{Food}: moringa oleifera
   \newline
    \textbf{Chemical}: MPG
    \newline
   
    \textbf{Food}: moringa oleifera
    \newline
   \textbf{Chemical}: phenolic glycoside
   \\\hline
      An unusual fatty acid, cis-9,cis-15-octadecadienoic acid, has been identified in the pulp lipids of mango (Mangifera indica L.) grown in the Philippines.
      &
         \textbf{Food}: mango
     \newline
    \textbf{Chemical}: cis-9,cis-15-octadecadienoic acid
   \newline
   
       \textbf{Food}: mango
     \newline
    \textbf{Chemical}: fatty acid
   \newline 
   
              \textbf{Food}: mangifera indica
     \newline
    \textbf{Chemical}: cis-9,cis-15-octadecadienoic acid
   \newline
   
       \textbf{Food}: mangifera indica
     \newline
    \textbf{Chemical}: fatty acid
         \\\hline
    (3,4-Dihydroxyphenyl)ethanol, commonly known as hydroxytyrosol (1), is the major phenolic antioxidant compound in olive oil, and it contributes to the beneficial properties of olive oil. &
         \textbf{Food}: olive oil
     \newline
    \textbf{Chemical}: (3,4-Dihydroxyphenyl)ethanol
   \newline
   
    \textbf{Food}: olive oil
     \newline
    \textbf{Chemical}: hydroxytyrosol
   \newline
   
     \textbf{Food}: olive oil
     \newline
    \textbf{Chemical}: phenolic
   \\\hline
   Two new phenolic acids forsythiayanoside C (1) and forsythiayanoside D (2), were isolated from the fruits of Forsythia suspense (Thunb.)
   &
   \textbf{Food}: fruits
     \newline
    \textbf{Chemical}: forsythiayanoside C
   \newline
   
   \textbf{Food}: fruits
     \newline
    \textbf{Chemical}: forsythiayanoside D
      \\\hline

    \end{tabular}
    \label{tab:re_examples_table}
\end{table}

\section{Conclusion}
In this paper, we present a Relation Extraction method for identifying the \textit{contains} relation between food and chemical entities in biomedical scientific literature, which is based on the fine-tuning the BERT, BioBERT and RoBERTa models. We evaluate the models trained on a small amount of ground truth, manually annotated data, and compare their performance to two scenarios where the same models are trained using an additional silver corpus annotated using a voting scheme of the tree models. The results indicate that in the majority of cases, such data augmentation has a positive impact on the models' performance. Out of the three evaluated models, the BioBERT model which uses unbalanced augmentation achieves the best results, with a macro averaged F1 score of 0.902.
The developed method can automatically suggest relations between food and chemical entities, which can then be checked by experts, and used to improve the coverage of existing food composition databases.

\section*{Acknowledgements}
This work has been supported by the Ad Futura grant for postgraduate study; the Slovenian Research Agency [research core funding programmes P2-0098]; the European Union's Horizon 2020 research and innovation programme [grant agreement 863059] (FNS-Cloud, Food Nutrition Security) and [grant agreement 101005259] (COMFOCUS); and the EFSA-funded project under [grant agreement GP/EFSA/AMU/2020/03/LOT2] (CAFETERIA).

\printbibliography
\end{document}